\documentclass[letterpaper, 10 pt, conference]{ieeeconf}  

\IEEEoverridecommandlockouts                              

\usepackage{graphics} 
\usepackage{graphicx}
\usepackage{epsfig} 
\usepackage{times} 
\usepackage{amsmath} 
\usepackage{amssymb}  

\usepackage{bm}
\usepackage[linesnumbered,ruled,vlined]{algorithm2e}
\usepackage[noend]{algpseudocode}
\usepackage[font={small}]{caption}
\usepackage{subcaption}
\usepackage{physics}
\usepackage{xcolor}
\usepackage{tikz}
\usetikzlibrary {arrows.meta}
\usetikzlibrary{fit,
                positioning}
\usepackage[belowskip=1ex]{subcaption}  

\usetikzlibrary{calc} 
\usetikzlibrary{positioning}
\usepackage{balance}
\usepackage{soul}
\usetikzlibrary{arrows}
\usepackage{cite}
\usepackage{url}
\usepackage{multirow}
\usepackage{booktabs} 
\usepackage[hidelinks]{hyperref}

\usepackage{amssymb}
\usepackage{amsfonts}
\usepackage{amsmath}
\usepackage{amsthm}
\usepackage{bm}

\usepackage[normalem]{ulem}                                        
\usepackage{marginnote}
\usepackage[textwidth=10ex,colorinlistoftodos]{todonotes}

\colorlet{mh}{red}
\colorlet{fwu}{red}
\colorlet{ywu}{blue}
\colorlet{kchen}{blue}
\colorlet{lchen}{green}
\colorlet{zbing}{green}
\colorlet{shaddadin}{purple}
\colorlet{iperez}{cyan}
\colorlet{schneider}{magenta}

\usepackage{amssymb}
\usepackage{mathtools}
\usepackage{sansmath}

\mathchardef\mhyphen="2D   

\newcommand{\RNum}[1]{\uppercase\expandafter{\romannumeral #1\relax}}

\usepackage{setspace}
\title{\LARGE \bf

Imitation-Guided Bimanual Planning for Stable Manipulation under Changing External Forces
}

\author{Kuanqi Cai$^{1,2*}$, Chunfeng Wang$^{3*}$, Zeqi Li$^{4}$, Haowen Yao$^{4}$, Weinan Chen$^{3}$, Luis Figueredo$^{5,4}$ \\ Aude Billard$^{2}$,~\IEEEmembership{Fellow,~IEEE}, Arash Ajoudani$^{1}$  
\thanks{$^{1}$HRI$^2$, Istituto Italiano Di Tecnologia (IIT), Genova, 16163, Italy.}
\thanks{$^{2}$LASA, School of Engineering, EPFL (Swiss Federal Institute of Technology in Lausanne), Lausanne, Switzerland. }
\thanks{$^{3}$Guangdong University of Technology, Guangzhou, 510006, China.}
\thanks{$^{4}$The Technical University of Munich, Munich, 85748, Germany.}
\thanks{$^{5}$The School of Computer Science, University of Nottingham, UK.}
\thanks{$^{*}$Equal contribution.}
\thanks{This work was supported by the European Union Horizon Project TORNADO (GA 101189557).}
}
\begin{document}

\maketitle
\thispagestyle{empty}
\pagestyle{empty}

\begin{abstract}
Robotic manipulation in dynamic environments often requires seamless transitions between different grasp types to maintain stability and efficiency. However, achieving smooth and adaptive grasp transitions remains a challenge, particularly when dealing with external forces and complex motion constraints. Existing grasp transition strategies often fail to account for varying external forces and do not optimize motion performance effectively. In this work, we propose an Imitation-Guided Bimanual Planning Framework that integrates efficient grasp transition strategies and motion performance optimization to enhance stability and dexterity in robotic manipulation. Our approach introduces Strategies for Sampling Stable Intersections in Grasp Manifolds for seamless transitions between uni-manual and bi-manual grasps, reducing computational costs and regrasping inefficiencies. Additionally, a Hierarchical Dual-Stage Motion Architecture combines an Imitation Learning-based Global Path Generator with a Quadratic Programming-driven Local Planner to ensure real-time motion feasibility, obstacle avoidance, and superior manipulability. The proposed method is evaluated through a series of force-intensive tasks, demonstrating significant improvements in grasp transition efficiency and motion performance. 
A video demonstrating our simulation results can be viewed at \href{https://youtu.be/3DhbUsv4eDo}{\textcolor{blue}{https://youtu.be/3DhbUsv4eDo}}.

\end{abstract}

\section{Introduction}

\begin{figure*}[t]
    \centering
    \includegraphics[width=17.5cm, height=9.5cm]{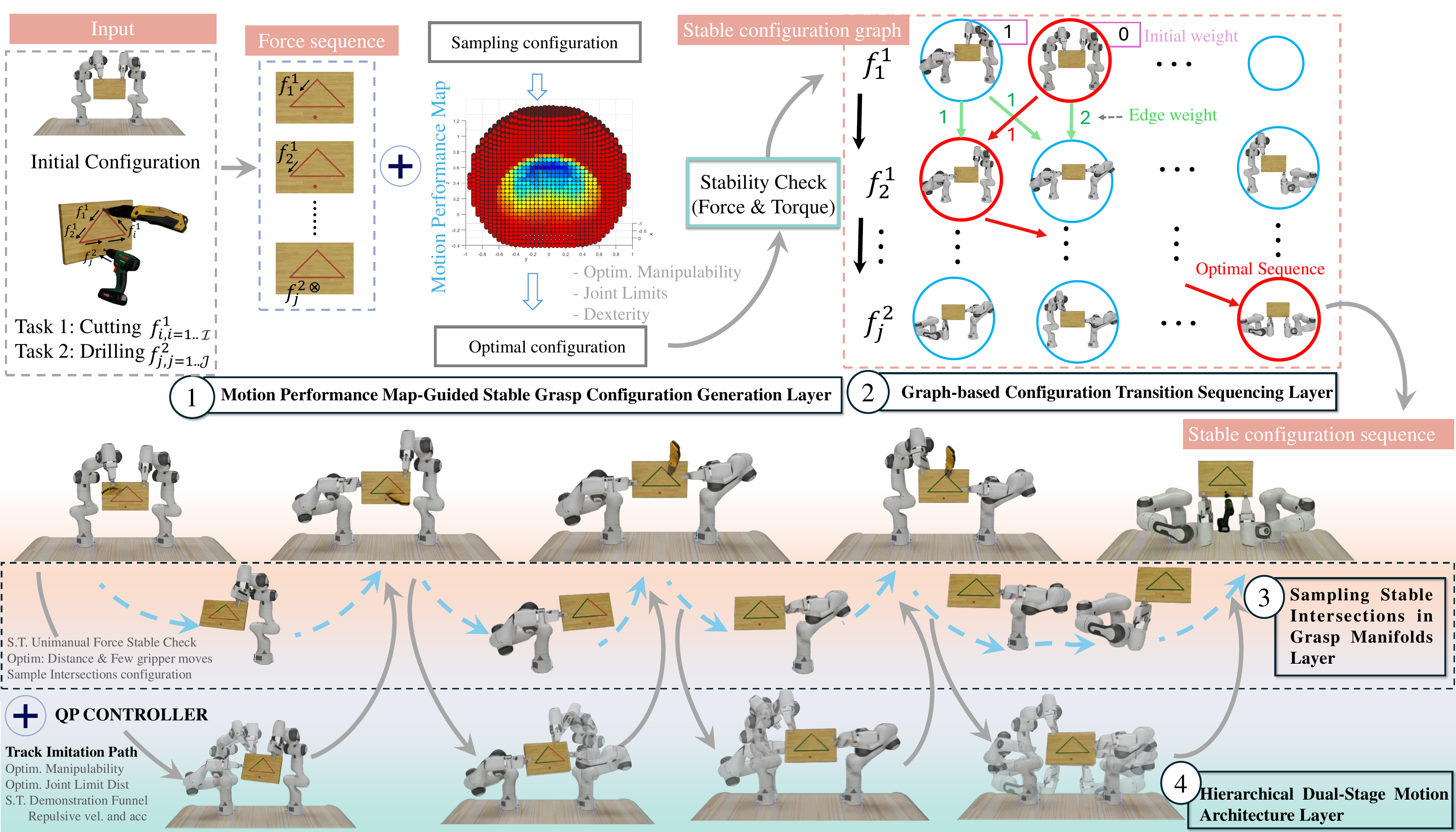}
    \caption{Overview of the proposed bimanual planning framework. This framework integrates unimanual and bimanual grasping strategies through four layers: motion performance map-guided stable grasp generation, graph-based configuration transition sequencing, sampling stable intersections in grasp manifolds, and a hierarchical dual-stage motion architecture. This framework ensures stable force equilibrium, efficient grasp transitions, and overall motion feasibility. In the stable configuration graph, edge weights represent the number of grasp transition steps. The initial weight is determined by the transition steps from the initial configuration to the first load-bearing configuration. In \textcircled{3}, blue dashed lines represent a bimanual plan transitioning through a single-arm configuration, while in \textcircled{4}, gray arrows indicate the actual execution.}
    \label{fig:framework}
\vspace{-10pt}
\end{figure*}

Robotic manipulation in dynamic forceful operations—such as collaborative cutting or drilling—demands real-time adaptation to varying external forces that critically affect grasp stability. Consider a human-robot woodworking scenario (Fig.~\ref{fig:framework}) where the robot must continuously adjust between uni-manual and bi-manual grasps to counteract changing cutting and drilling forces. This fundamental requirement exposes two unresolved challenges in existing methods: efficient grasp transitions by minimizing execution time and arm movement and motion performance awareness, as crucial metrics like manipulability and joint limits essential for control safety are often overlooked. To bridge this gap, we propose an imitation-guided planning framework that integrates efficient grasp transitions with motion performance constraints, ensuring both stability and dexterity in forceful tasks.

Multi-step manipulation planners have long tackled regrasping and grasping transitions~\cite{tournassoud1987regrasping}. Traditional grasping involves transporting an object by repeatedly releasing and regrasping it as needed~\cite{ma2018regrasp}. Conventional regrasp planners rely on a supporting surface for single-arm manipulation~\cite{wanwwsingle, Rohrdanz_single}, while recent research extends these strategies to dual-arm scenarios~\cite{cruciani2019dual, qindualarm, wan2016developing}.
However, existing methods do not explicitly account for dynamic external forces, which vary over time, nor do they optimize regrasp transitions during forceful interactions. Studies in forceful human-robot collaboration~\cite{kosuge1997control,rozo2016learning,abi2017learning} focus on regulating contact forces but assume a fixed or pre-determined grasp. The key challenge remains: determining where and how a robot should grasp for stability and when to transition seamlessly under complex external forces. Recent works~\cite{chen2019manipulation,cheng2018etc,chen2020manipulation} have made progress, but achieving stable grasps that withstand varying forces while ensuring efficient planning, manipulability, and dexterity remains difficult.
This paper addresses two key challenges in forceful robotic manipulation: \textbf{efficient grasp transitions} and \textbf{motion performance optimization}, proposing a novel framework to overcome them.

\textbf{Efficient Grasp Transitions.} Previous methods~\cite{chen2019manipulation, chen2020manipulation} mostly rely on random sampling-based planners for grasp transitions, often resulting in high computational costs and unstable changes. To reduce task execution time and minimize the movement distance of the robot arm, we introduce \textit{Strategies for Sampling Stable Intersections in Grasp Manifolds} for seamless uni-manual and bi-manual transitions. Our \textit{Directional Gradient-Based Resampling} locally adjusts the unimanual manipulator along the negative gradient, ensuring stability while maintaining a secure unimanual grasp and minimizing movement. For tasks with multiple grasp changes, \textit{Multi-Grasp Transition Check (MTC)} identifies a shared intermediate configuration, reducing redundant regrasping. To further boost efficiency, we propose a \textit{Hierarchical Dual-Stage Motion Architecture}, combining an \textit{Imitation Learning-based Global Path Generator} with a QP-driven local planner for real-time motion optimization and obstacle avoidance, enabling faster collision-free path generation than sampling-based methods.

\textbf{Motion Performance Optimization.} 
In forceful operations, robotic stability is determined by three motion performance factors: manipulability, dexterity, and joint limits, which ensure kinematic feasibility under varying forces. However, many recent methods~\cite{raessa2020human,figueredo2021planning} tend to overlook these critical aspects.
Our framework optimizes motion performance at both the grasp configuration level and during execution. To enhance manipulability and avoid kinematic limitations, we introduce a \textit{Motion Performance Map} that encodes feasibility, manipulability, and joint limit proximity. This map guides grasp sampling toward optimal workspace regions, improving selection efficiency. During execution, we enforce manipulability constraints within the QP framework, ensuring control authority over the end-effector while avoiding singularities and joint limits. This real-time optimization enables stable, collision-free motion.

By integrating these advancements, our framework enhances grasp transition efficiency while ensuring superior motion performance in both grasp selection and execution, effectively addressing key limitations in existing forceful robotic planning approaches.

\section{Problem Definition}
This work introduces an imitation-guided hierarchical planning framework that enables efficient stability-aware grasp transitions and constrained motion optimization for real-time adaptation to dynamic external forces in forceful robotic manipulation

\subsection{Preliminaries}

We define the force vector as \(\bm{F} = (\bm{f}, \bm{p})\), where \( \bm{f} \in \mathbb{R}^{6} \) is a generalized force (force/torque) applied at a pose \( \bm{p} \in \mathrm{SE}(3) \). Any forceful task can be represented as a sequence of forces, \(\bm{\mathcal{F}} = (\bm{F_h})_{h=1}^m\), with continuous tasks discretized into force sequences. For example, a cutting task on a triangular shape involves $\mathcal{I}$ force vectors at varying poses \( \bm{p} \), while a drilling task is discretized into $\mathcal{J}$ force vectors at a fixed location, as shown in Fig.~\ref{fig:framework}.  
Consider a dual-arm robotic platform, each arm equipped with a gripper. Let \(\mathrm{SE}(3)\) be the object configuration space, \(\bm{C^L} \) and \( \bm{C^R} \) the joint configuration spaces of the left and right manipulators, respectively. The system's composite configuration space is $\bm{\mathcal{C}} = \bm{C^L} \times \bm{C^R} \times \mathrm{SE}(3)$, where any bi-manual configuration \( \bm{q} \in \bm{\mathcal{C}} \) is expressed as $\bm{q} = (\bm{q^L}, \bm{q^R}, \bm{x})$ with \( \bm{q^L} \in \bm{C^L} \), \( \bm{q^R} \in \bm{C^R} \), and \( \bm{x} \in \mathrm{SE}(3) \) denoting the object's Cartesian pose. Unimanual configurations are written as \( \bm{q} = (\bm{q^L}, \bm{x}) \) or \( (\bm{q^R}, \bm{x}) \).  
Applying forward kinematics, the composite configuration \( \bm{q} \) maps to its corresponding grasp configuration \( \bm{g} \)\textcolor{black}{$\in \mathbb{R}^{12}$}. 
The configuration space \(\bm{\mathcal{C}}\) is represented as a union of lower-dimensional manifolds: $\bm{\mathcal{C}} = \bigcup_{\bm{g} \in \Gamma} M(\bm{g})$, where \(\bm{\Gamma}\) is the set of all possible unimanual and bimanual grasps, and each manifold \( M(\bm{g}) \) corresponds to a specific grasp \(\bm{g} \). Thus, transitioning between unimanual and bimanual grasps equates to moving across these manifolds.

\subsection{Overview of the problem}
Robotic manipulation tasks involving forceful interactions, such as cutting, drilling, or assembling, require stable grasp configurations to ensure both stability and efficiency. Given an ordered sequence of force operations \( (\bm{F}_p)_{p=1}^m \) on a target object, the goal is to determine a minimal yet feasible sequence of grasp configurations for stable and efficient execution while maintaining motion constraints.  
This paper proposes an optimized strategy to identify a minimal set of stable grasp configurations $\bm{\mathcal{Q}'} = \{ \bm{q}_j \}_{j=1}^n \subseteq \bm{\mathcal{C}}$ where each \( \bm{q}_j \) accommodates a contiguous subset of force operations $\bm{\mathcal{F}_j} = \{ \bm{F_h} \}_{h=a_j}^{b_j} \subseteq \bm{\mathcal{F}}$
ensuring stability while minimizing transitions. The sequence \( \bm{\mathcal{Q}}' \) fully covers all force operations in order, satisfying $\bm{\mathcal{F}} = \bigcup_{j=1}^n \mathcal{F_j}$.  

Beyond optimizing grasp configurations, real-time transition planning is crucial. Each transition from \( \bm{q_{j-1}} \) to \( \bm{q_j} \) requires a feasible trajectory  $\mathbf{t_j}$ that ensures collision avoidance, manipulability, and joint limit compliance. Efficient, collision-free transitions are essential for reliable execution.

\noindent\textbf{Problem Statement}:  
Given an ordered sequence of force operations \( (\bm{F_h})_{h=1}^m \), the goal is to determine a minimal set of stable grasp configurations $\bm{\mathcal{Q}'} = \{ \bm{q_j} \}_{j=1}^n$ that cover all force operations in \( \bm{\mathcal{F}} \) while preserving their order. Each \( \bm{q_j} \) must satisfy motion constraints and ensure system stability under its associated force subset \( \bm{\mathcal{F}_j} \). Transitions between consecutive configurations \( \bm{q_{j-1}} \) and \( \bm{q_j} \) must follow a collision-free, high-performance trajectory $\mathbf{t}_j$. The objective is to minimize \( |\bm{\mathcal{Q}'}| \), reducing unnecessary grasp transitions while maintaining force equilibrium, motion constraints, and efficient movement in terms of cost and time.

\section{Method}

\subsection{Method Overview}

To address the above problem, we propose a hierarchical planning framework that integrates unimanual and bimanual grasping to optimize grasp transitions while ensuring stability and good motion performance. The framework comprises four layers:
1. \textit{Motion Performance Map-Guided Stable Grasp Configuration Generation};
2. \textit{Graph-Based Grasp Transition Sequencing};
3. \textit{Strategies for Sampling Stable Intersections in Grasp Manifolds Layer};
4. \textit{Hierarchical Dual-Stage Motion Architecture}. 

As shown in Fig.~\ref{fig:framework} and detailed in Alg.~\ref{alg:IGB}, the framework takes the robot's initial configuration and external force vectors as input. It constructs a motion performance map (line1 of Alg.\ref{alg:IGB}) to identify optimal grasp configurations within the object's graspable region (e.g., the edge of a wooden board), ensuring high manipulability, dexterity, and joint-limit avoidance.  
Next, force-torque balance checks (Alg.\ref{alg:IGB}, lines 3-4) determine stable grasp candidates, forming a \emph{Stable Configurations Graph} whose nodes represent feasible grasps and edges encode regrasp transitions. A graph search (lines~5-6) then seeks the path following Djikstra while guaranteeing force equilibrium.
Building on these stable configurations, the \emph{Sampling Stable Intersections in Grasp Manifolds} layer (Alg.\ref{alg:IGB}, lines 7-9) refines unimanual configurations to balance gravitational forces and shorten trajectories, reducing intermediates. Once stable intersections are identified, an imitation learning-based motion generator (Alg.\ref{alg:IGB}, line 9) computes smooth trajectories as the global path in the hierarchical motion architecture. Finally, a QP controller (Alg.~\ref{alg:IGB}, line 10) optimizes motion performance and obstacle avoidance for reliable execution. The following sections detail each component.

\begin{algorithm}[t]
\caption{Imitation-Guided Bimanual Planner}
\label{alg:IGB}
\DontPrintSemicolon
\KwIn{Forces: $\bm{F}=\{(\bm{f}_i,\bm{p}_i)\}_{i=1}^m$, Initial: $\bm{q}_0$}
\KwOut{Joint Velocity: $\dot{\bm{\mathfrak{q}}}$}

$\omega(\bm{q_k})\leftarrow{BuildMap}()$ $\rhd$ Eqs. (2) - (5)\\
${\bm{g}*}\leftarrow$ $OptimailGraspConfiguration$($\omega$) \\
\ForEach{$(\bm{f}_i,\bm{p}_i)\in \bm{F}$}{
  $\bm{\mathcal{Q}_\text{stable}}\leftarrow ForceCheck({\bm{g}*})$ $\rhd$ Eqs. (6) - (9)
}
$(Graph, Cost)\leftarrow BuildGraph \bigl(\bm{\mathcal{Q}_\text{stable}})$ $\rhd$ Eq. (10) \\
$\bm{Q}\leftarrow GraphSearch \bigl(Graph,\bm{q_0})$ $\rhd$ Eq. (11)\\
\ForEach{consecutive $(\bm{g}_s,\bm{g}_t)\in \bm{Q}$}{
  $\bm{g_{ic}^{new}}\leftarrow Sampling Stable Intersections\bigl(\bm{g_s},\bm{g_t})$ $\rhd$ Eqs. (12) - (15)\\
  $\bm{Q^*} \leftarrow \bm{g_{ic}^{new}}$
}
$\text{GlobalPath}\leftarrow ImitationPath(Q^*,\text{demoPath})$ $\rhd$ Eqs. (16) - (20) \\
$\dot{\bm{\mathfrak{q}}}\leftarrow QPController(\text{GlobalnPath}, \text{Constraints})$ $\rhd$ Eqs. (21) - (23)\\
\Return{$\dot{\bm{\mathfrak{q}}}$}
\end{algorithm}

\subsection{Motion Performance Map-Driven Stable Grasp Configuration Generation Layer}
\label{map}
\begin{figure*}
    \centering
    \includegraphics[width=0.95\linewidth, height=3.5cm]{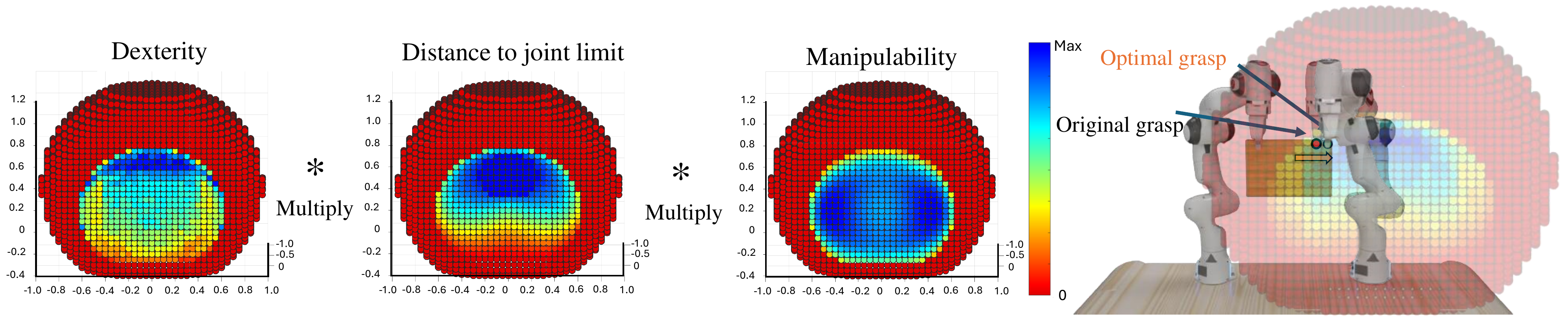}
    \caption{The voxelized workspace of the Franka Emika robot integrates motion performance data, incorporating dexterity, joint limit proximity, and manipulability to optimize grasp configuration selection, enhancing kinematic flexibility.}
    \label{fig:map}
\end{figure*}

This layer optimizes motion configurations to maintain object stability under force \( \bm{F} \) at position \( \bm{p} \) while enhancing \textcolor{black}{manipulability}, dexterity, and joint-limit avoidance, ensuring optimal grasp selection for manipulability and safety.

\subsubsection{Formulation of the Motion Performance Map and Score Functions}
To generate the motion performance map, we discretize the end-effector (EE) workspace in both position and orientation. The Cartesian space is sampled as \( \bm{\Omega_{\text{pos}}}  \in \mathbb{R}^3\) with a fixed resolution (0.1m in this study), while orientation is discretized into \( \bm{\Omega_{\text{dir}}}  \in \mathbb{R}^3
  \) (principal EE \( z \)-axis directions) and \( \bm{\Omega_{\text{rot}}}  \in \mathbb{R}^3
 \) (rotations around each direction). The complete voxelized set is:  
\begin{equation}
\bm{V} = \{ \bm{v_k} = (\bm{p_k}, \bm{d_k}, \bm{\phi_k}) \mid \bm{p_k} {\in} \bm{\Omega_{\text{pos}}}, \bm{d_k} {\in} \bm{\Omega_{\text{dir}}}, \bm{\phi_k} {\in} \bm{\Omega_{\text{rot}}} \}.
\end{equation}
Each voxel \( \bm{v_k} \) represents a candidate EE pose \( \bm{T_{\text{EE}}(v_k)} \). The motion performance score \( \omega(\cdot) \) evaluates grasp configuration \( \bm{g} \in \mathrm{SE}(3)\) based on (a) manipulability, (b) dexterity, and (c) joint-limit proximity. Given a feasible joint configuration \( \bm{q_k} \) from \( \bm{v_k} \), manipulability is defined as:  
\begin{equation}
m(\bm{q_k}) = \sqrt{\det(\bm{J} \bm{J}^\top)},
\end{equation}
where \( \bm{J} \) is the geometric Jacobian. A higher \( m(\cdot) \) value indicates better manipulability, facilitating smooth movement and avoiding singularities.  

Dexterity is quantified by the number of feasible nullspace solutions within the discretized swivel angle space~\cite{haowenEDM}, \textcolor{black}{which is the rotation of the arm’s upper–lower plane about the fixed shoulder–wrist axis.} 
\begin{equation}
\eta(\bm{q_k}) = \left| \{ \theta_{\text{swivel}}(\bm{q_k}) \mid \theta_{\text{swivel}}(\bm{q_k}) \in \bm{\Theta_{\text{free}}} \} \right|,
\end{equation}
where \( \theta_{\text{swivel}}(\bm{q_k}) \) denotes the EE’s feasible swivel angle at configuration \( \bm{q_k} \), and \( \bm{\Theta_{\text{free}}} \) is the predefined range of allowable swivel angles, ensuring compliance with physical and task constraints.  

To maintain joint positions away from limits, we define a joint-limit proximity factor~\cite{vahrenkamp2012manipulability}:  
\begin{equation}
l(\bm{q_k}) = 1 - \exp\left( -k \prod_{j=1}^{n} \frac{(\bm{q^j_k} -  \bm{\underline{q}^j_{lim}})(\bm{\overline{q}^j_{lim}} - \bm{q^j_k})}{(\bm{\overline{q}^j_{lim}} - \bm{\underline{q}^j_{lim})}^2} \right),
\end{equation}
where \( \bm{\underline{q}^j_{lim}} \) and \( \bm{\overline{q}^j_{lim}} \) are joint position bounds, and $k \in \mathbb{R_+}$ is a positive weight constant. $n$ represents the number of degrees of freedom (DOF) of the manipulator.
We take the median value of each factor across viable configurations, normalize them over voxelized set \( V \), and compute the \textcolor{black}{motion performance score}:  
\begin{equation}
\omega(\bm{q_k}) = \hat{m}(\bm{q_k}) \cdot \hat{\eta}(\bm{q_k}) \cdot \hat{l}(\bm{q_k}).
\end{equation}
The constructed map (Fig.~\ref{fig:map}) guides the selection of high-scoring grasp configurations within the object's graspable region, which are then validated for force equilibrium.

\subsubsection{Force Stability Check With Bimanual configuration}

We formulate forceful operations mathematically and describe how the planner assesses the force stability of candidate grasp configurations under a given force \( \bm{f} \). The stability check ensures that a grasp configuration, using both arms, can resist the applied force by evaluating two key constraints: grasp-wrench limits and joint torque limits.

The total wrench exerted by both grippers must counteract the external force \( \bm{f'} \)\footnote{Deviations in external force can be handled using the Conic Model~\cite{chen2020manipulation}.}, \textcolor{black}{including gravity and other environmental interactions,} mapped to a common reference frame via \( R(\bm{p}) \). The equilibrium condition is:
\begin{equation}
    \begin{aligned}
        & \bm{W} \bm{f}' + \sum_{i \in \{\text{left}, \text{right}\}} \bm{J}_i^{\top} \bm{f_{g,i}} = 0,  \quad \bm{f'} = R(p)\bm{f}, \\
        & \bm{f_{g,\min}} \leq \bm{f_{g,i}} \leq \bm{f_{g,\max}}, \quad \forall i \in \{\text{left, right}\},
    \end{aligned}
    \label{eq:force_check1}
\end{equation}
where \( \bm{W} \in \mathbb{R}^{6 \times 6m} \) is the grasp matrix~\cite{borst2004grasp}, mapping gripper forces to the object's resultant wrench, \( \bm{J_i} \in \mathbb{R}^{6 \times n_i} \) is the Jacobian matrix, and \( \bm{f_{g,i}} \in \mathbb{R}^{6} \) is the force applied by each gripper. Each gripper's force must lie within its resistible wrench space~\cite{mishra1987existence}, approximated by a 6D axis-aligned bounding box. \( \bm{f_{g,\max}} \) and \( \bm{f_{g,\min}} \) define the upper and lower grasp force limits, which are determined using friction cone constraints or experimental measurements. 

The robot must ensure that the required joint torques remain within allowable limits. The joint torques for both arms must remain within allowable limits:
\begin{equation}
    \begin{aligned}
        & \bm{\tau}_i = \bm{J}_i^\top \bm{f_{g,i}}, \quad \forall i \in \{\text{left, right}\}, \\
        & \bm{\tau_{\min}} \leq \bm{\tau_i} \leq \bm{\tau_{\max}}.
    \end{aligned}
    \label{eq:force_check2}
\end{equation}
Here, \( \bm{\tau_i} \in \mathbb{R}^{n_i} \) represents the joint torque vector for each arm, while \( \bm{\tau_{\min}} \) and \( \bm{\tau_{\max}} \) denote the lower and upper torque limits, ensuring the actuators operate within safe ranges.

By enforcing these constraints, we identify grasp configurations that can support external forces effectively.

\subsection{Graph-based Grasp Transition Sequencing Layer}
\label{grasp_sample}

After identifying all feasible configurations \( \bm{q} \) through stability checks for each forceful operation \( \bm{F}_h \) in \( \{\bm{F}_h\}_{h=1}^m \), we construct a directed acyclic weighted graph—\textit{stable configuration graph} \( \bm{G_0} \)—to represent the regrasping process. Specifically, let \( \bm{Q^i}\) be the set of stable configurations resisting operation \( \bm{f^i_h} \), forming the first row of the graph in Fig.~\ref{fig:framework}. Each configuration in \(\bm{ Q^i} \) is a node in the \( i \)-th column. If two nodes \( \bm{q^i} \in \bm{Q^i} \) and \( \bm{q^{i+1}} \in \bm{Q^{i+1}} \) in adjacent rows can transition via \( r \) regrasp actions (\( r = 0, 1, 2, \dots \)), we add a directed edge from \( \bm{q^i} \) to \( \bm{q^{i+1}} \) weighted by the number of regrasp actions required:
\begin{equation}
    c\bigl(\bm{q^i}, \bm{q^{i+1}}\bigr) = r.
\end{equation}
For example, \( r = 0 \) if no regrasp is needed, \( r = 1 \) if one arm must regrasp, and \( r = 2 \) if both arms must switch grasp poses. Consequently, \( \bm{G_0} \) remains a directed acyclic graph where each row corresponds to an operation \( \bm{f_i} \), nodes represent stable configurations, and edge weights indicate regrasp costs. The objective is to find a path minimizing the total regrasp cost, formulated as:
\begin{equation}
    \min_{\{\bm{q^1},\dots,\bm{q^m}\}} \sum_{i=1}^{m} c\bigl(\bm{q^i}, \bm{q^{i+1}}\bigr) + c(\bm{q^{\text{init}}}, \bm{q^1}).
\end{equation}
Here, \( c(\bm{q^{\text{init}}}, \bm{q^1}) \) represents the cost from the initial configuration \( \bm{q^{\text{init}}} \) to \( \bm{q^1} \). If all operations \( \{\bm{F_h}\} \) are known beforehand, the full graph can be constructed, and standard shortest-path algorithms (e.g., Dijkstra’s) can find the optimal sequence. If operations arrive incrementally, the graph expands dynamically, and a greedy search provides a feasible sequence with minimal grasp transitions, following our prior work~\cite{chen2019manipulation, chen2020manipulation}.

\begin{figure}
    \centering
    \includegraphics[trim=5 1 5 1, clip, width=\linewidth]{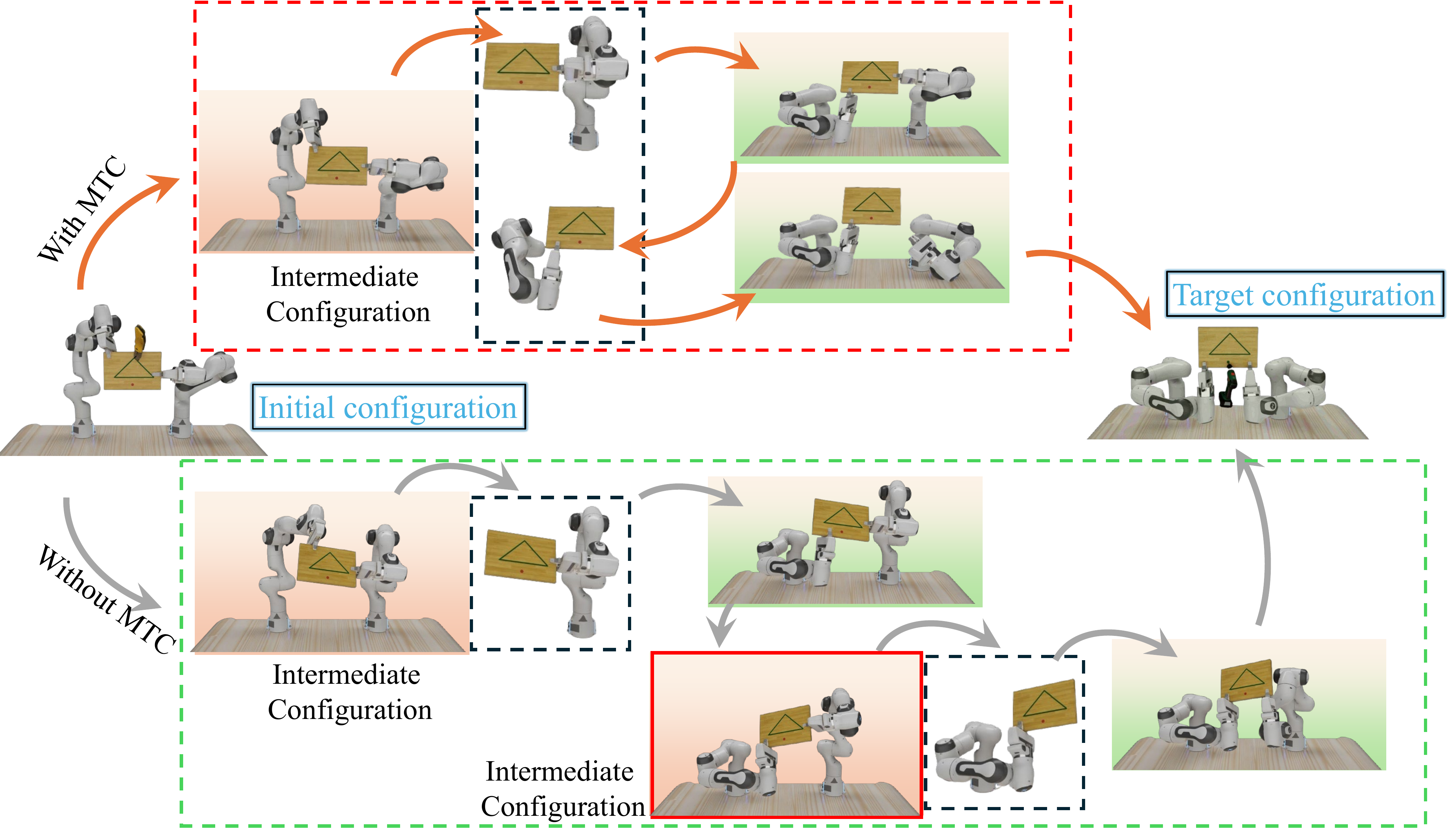}
    \caption{A diagram illustrating the grasp transition process w/ and w/o the \textit{Multi-Grasp Transition Check} (MTC) method. The black dashed box indicates intermediate configurations supporting single-arm board holding, while the red solid box marks the object configuration needed without the MTC method.}
    \label{fig:mtc}
\end{figure}
\subsection{Strategies for Sampling Stable Intersections in Grasp Manifolds Layer}
After determining a series of bimanual grasp configurations, this section focuses on identifying unimanual grasps that enable smooth transitions while ensuring gravitational stability. To achieve this, we introduce Sampling Stable Intersections in Grasp Manifolds, optimizing unimanual grasp selection.
During the transition from bimanual to unimanual grasping, both arms reposition the object to an intermediate configuration where one arm fully supports its weight, allowing the other to release and regrasp. The initial intermediate configuration is sampled based on a force balance strategy, with $i$ selecting either the left or right gripper to achieve equilibrium as defined in Eq.~\eqref{eq:force_check1}-\eqref{eq:force_check2}.
To ensure a seamless and efficient transition from the initial grasp $\bm{g_b}$ to the intermediate configuration $\bm{g_i}$, we propose a Directional Gradient-Based Resampling strategy, which minimizes displacement between $\bm{g_b}$ and $\bm{g_{ic}}$. The objective function governing this minimization is defined as:
\begin{equation}
D(\bm{g}) = \frac{1}{2} \| \bm{g} - \bm{g_b} \|^2.
\label{eq:motion1}
\end{equation}
Accordingly, an update along the negative gradient direction is given by:
\begin{equation}
\bm{g^{\text{new}}_{ic}} = \bm{g_{ic}} - \eta \nabla D(\bm{g})  + r,
\end{equation}
where $\eta > 0$ represents the step size. To mitigate the impact of uncertainties and circumvent local minima, we incorporate a stochastic perturbation term $r \sim \mathcal{N}(0, \sigma^2 \bm{I})$.
Subsequently, the candidate solution is projected onto the grasp manifold set $\mathcal{C}$, ensuring feasibility by verifying inverse kinematics and stability constraints:
\begin{equation}
\bm{g^{opt}_{ic}} = \Pi_\mathcal{C} \left( \bm{g^{new}_{ic}} \right).
\label{eq:project}
\end{equation}
Through this method, the system efficiently identifies the optimal unimanual transition grasp that ensures gravitational stability while minimizing movement distance.

\noindent{\textbf{Multi-Grasp Transition Check (MTC)}}: When multiple grasp transitions are needed (e.g., two transitions in Fig.~\ref{fig:mtc}), using intermediate configurations that support only one arm (black dotted box) forces separate transitions for each grasp, increasing movement and reducing efficiency (green dashed box).
To address this, we propose the MTC, which selects a single intermediate configuration that simultaneously supports multiple unimanual grasps (red dashed box in Fig.~\ref{fig:mtc}). By consolidating multiple grasps into one configuration, MTC minimizes transitions and improves efficiency.
On the grasp manifold \( \mathcal{C} \), the optimization maximizes stable grasps while minimizing displacement:
\begin{equation}
    \bm{g^{\text{new}}_{ic}} = \arg\max_{\bm{g} \in \bm{\mathcal{C}}} \left[ \sum_{i=1}^{N} \delta_i(\bm{g}) - \lambda \| \bm{g} - \bm{g_b} \|^2 \right],
\label{eq:Motion1}
\end{equation}
where \( \lambda > 0 \) balances grasp stability and displacement. A configuration may be discarded if excessive displacement is required, prioritizing closer configurations with more transitions.  
\( \delta_i(\bm{g}) \) is an indicator function denoting whether the \( i \)-th grasp is stable under configuration \( \bm{g} \). All configurations must pass the force equilibrium evaluate before being considered.

MTC optimizes configurations to enhance stability and minimize unnecessary movements, improving multi-grasp transition efficiency. The optimal intermediate configurations \( \bm{g^{\text{opt}}_{ic}} \), derived from Eq.~\eqref{eq:project}, satisfy inverse kinematics and stability constraints.

\begin{figure*}[htp]
    \centering
    \subfloat[Random Cutting]{
        \includegraphics[width=0.21\linewidth, height=2.3cm]{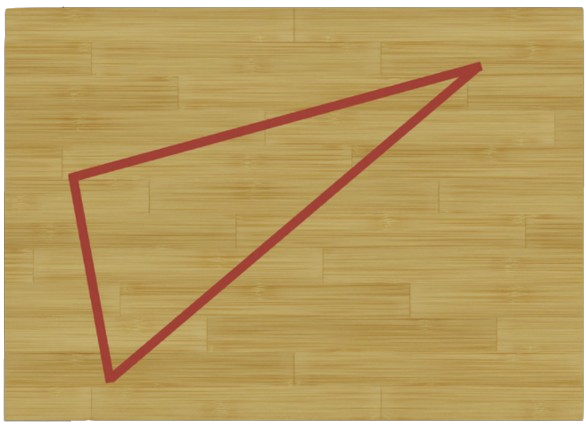}
    }
    \subfloat[Random Drilling (Case 1)]{
        \includegraphics[width=0.21\linewidth, height=2.3cm]{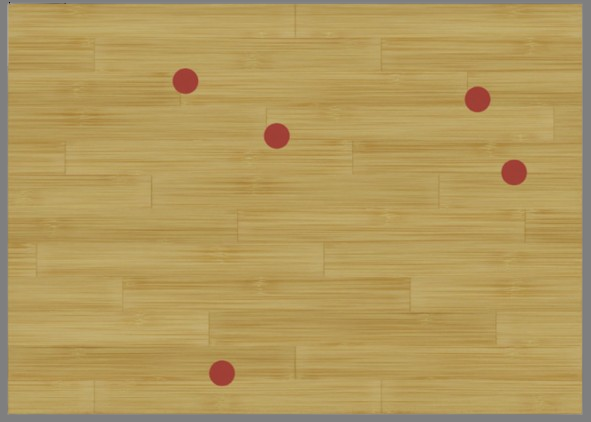}
    }
    \subfloat[Random Drilling (Case 2)]{
        \includegraphics[width=0.21\linewidth, height=2.3cm]{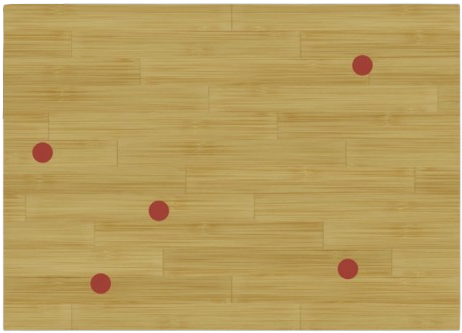}
    }
    \subfloat[Combined Cutting and Drilling]{
        \includegraphics[width=0.21\linewidth, height=2.3cm]{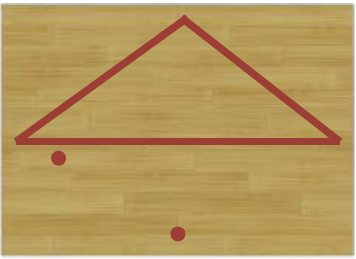}
    } 
    \caption{Visualization of Four External Force Variation Task Scenarios.}
    \label{fig:tasks}
\end{figure*}

\begin{figure}[htp]
\centering  
\includegraphics[width=0.48\textwidth, height=0.23\textheight]{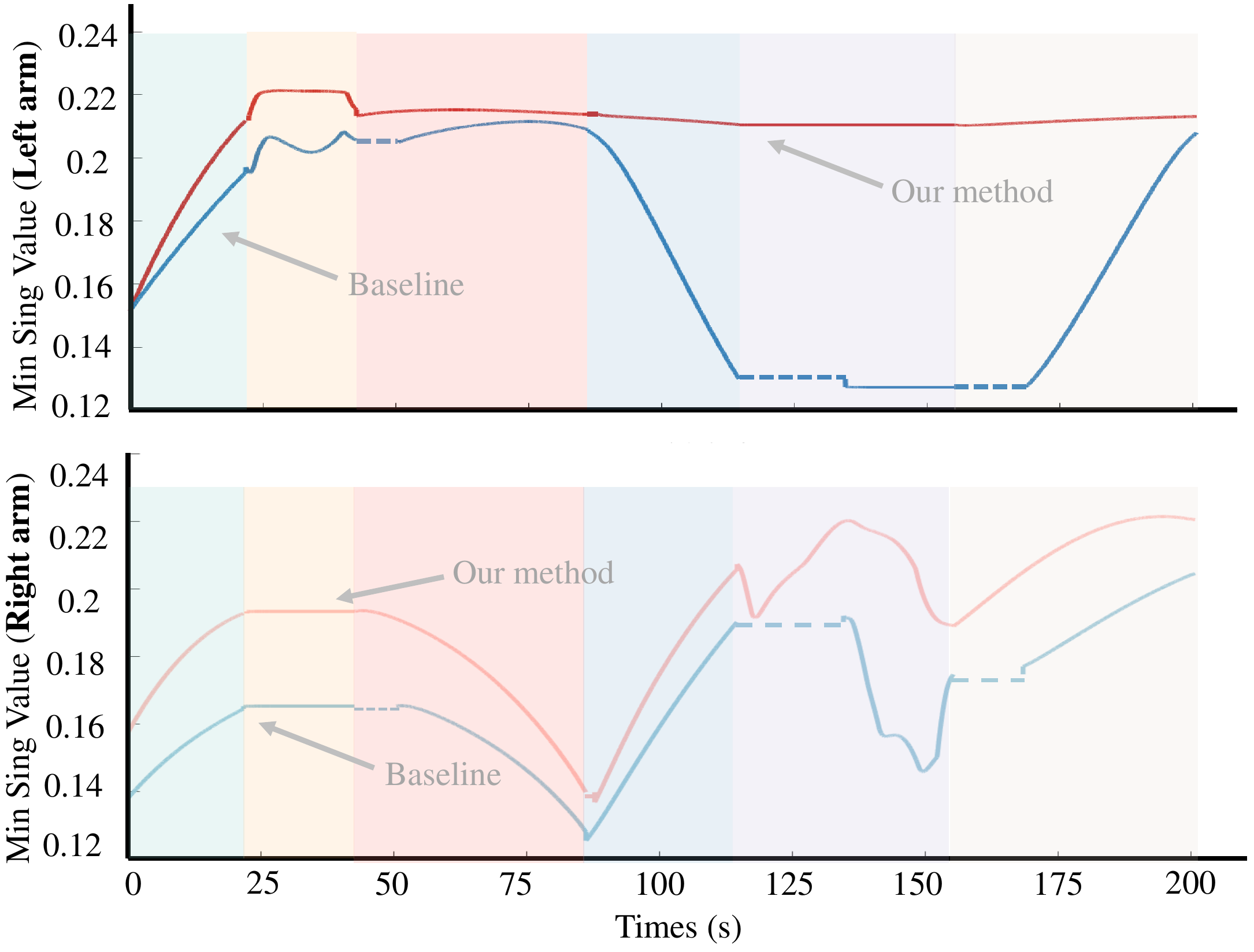}
\caption {The comparison between our planner and baseline. Different colored regions represent different grasp transitions.} 
\label{fig:exp_sing} 
\end{figure}

\subsection{Hierarchical Dual-Stage Motion Architecture Layer}
\label{qp}
In this layer, after obtaining a sequence of stable configurations—including both bimanual and unimanual states—we introduce a hierarchical dual-stage motion architecture to ensure efficient and safe transitions for the dual-arm robot.
On the global stage, we propose an \textbf{imitation learning-based global path generator} that adapts robot trajectories from multiple human demonstrations. By capturing motion variability, we compute the mean and variance, enabling flexible task reproduction. Each end-effector demonstration trajectory, \( \bm{\mathbf{x}^D} \), is provided, and the corresponding geometric primitive \( \bm{\mathbf{x}^P} \) is obtained by fitting multiple demonstrations:
\begin{equation} 
\bm{\mathbf{x}^P} = \bm{\Psi}^\top \bm{w} + \epsilon_y,
\end{equation}
where \( \bm{\Psi} \) is a set of Gaussian basis functions, and \( \epsilon_y \sim \mathcal{N}(0, \bm{\Sigma_y}) \) is zero-mean Gaussian noise with covariance \( \bm{\Sigma_y} \). The weight vector \( \bm{w} \) follows a Gaussian distribution~\cite{paraschos2015model}:
\begin{equation}
p(\bm{w}) = \mathcal{N}(\bm{\mu_w}, \bm{\Sigma_w}).
\end{equation}
By learning from multiple trajectories, the robot tracks the mean trajectory \( \bm{\overline{\mathbf{x}}^D} \) while leveraging variance \( \bm{\Sigma} \) for adaptability:
\begin{equation}
\bm{\overline{\mathbf{x}}^D} = \bm{\Psi}^\top \bm{\mu_w}, \quad \bm{\Sigma} = \bm{\Psi}^\top \bm{\Sigma_w} \bm{\Psi} + \bm{\Sigma_y}. 
\end{equation}
This approach derives a desired trajectory and covariance from multiple demonstrations, following our previous work~\cite{caid2a} and aligning with the principles of the Probabilistic Movement Primitives (ProMPs) framework~\cite{paraschos2015model}.

For trajectory reproduction towards a new goal, we first project $\bm{\overline{\mathbf{x}}^D}$ into the dual quaternion space\footnote{The dual quaternion space provides a compact, singularity-free representation for rigid body transformations by coupling rotation and translation in $\text{Spin}(3) \ltimes \mathbb{R}^3$, enabling efficient interpolation compared to traditional $\text{SE}(3)$ methods. See~\cite{selig2007geometric} for details.} $\bm{\overline{\mathfrak{x}}^D}$. Subsequently, we compute the incremental object-centric transformations between successive poses within $\bm{\overline{\mathfrak{x}}^D}$.
\begin{equation}
\bm{\delta^D_{i}} = \left(\bm{\overline{\mathfrak{x}}^D_{i-1}}\right)^{-1} \cdot \bm{\overline{\mathfrak{x}}^D_{\eta}}, \quad i = 2, \ldots, \eta.
\end{equation}
Given a new goal pose ${\mathfrak{x}}^N_f$, the imitation path $\{\bm{\overline{\mathfrak{x}}^{I}_i}\}^{\eta}_{i=1}$ is generated by applying the transformation to the new goal:
\begin{equation}
\bm{\overline{\mathfrak{x}}^I_{i}} = \bm{{\mathfrak{x}}^N_f} \cdot \bm{\delta^D_{i}}, \quad i = 2, \ldots, \eta.
\end{equation}
To ensure smooth transitions between the initial pose $\bm{\mathbf{q}_s}$ and the new goal $\bm{\mathbf{q}_g}$, we employ the screw linear interpolation method ($C^1-ScLERP$)~\cite{allmendinger2018coordinate}.
The proposed generator guarantees that the reproduced global trajectories respect both demonstrated constraints and maintain smooth transitions.

On the \textit{local stage}, we employ the Quadratic Programming formulation to integrate obstacle avoidance, joint‐limit compliance, slack, and manipulability in real-time to track the global trajectory.
We define the following cost function:
\begin{equation}
\begin{aligned}
\min_{\bm{u}} \Big( 
& \, \kappa_1 \, \| \bm{J} \bm{u} - \gamma(\bm{\overline{\mathfrak{x}}^I_{new}}) \|^2 
+ \kappa_2 \, \| \mathbf{J}_\sigma \bm{u} \|^2 
+ \kappa_3 \, \| \mathbf{J}_s \bm{u} \|^2 
\Big) \\[6pt]
\text{s.t.} \quad 
& \, \mathbf{A}_s \Delta t \cdot \bm{u} \leq \mathbf{b}_s, \ \ \mathbf{lb}_p \leq \bm{u} \leq \mathbf{ub}_p, \\ 
& \, \mathbf{lb}_v \leq \bm{u} \leq \mathbf{ub}_v, \ \  \mathbf{lb}_a \leq \bm{u} \leq \mathbf{ub}_a,
\end{aligned}
\end{equation}
where $\bm{u}=[\dot{\bm{\mathfrak{q}}}^T, \, \bm{\mathbf{s}}^T]^T$ is the optimization variable. $\bm{\mathfrak{q}} \in \mathbb{R}^{r}$ denote the vector of joint angles, and  $\dot{\bm{\mathfrak{q}}} \in \mathbb{R}^{r}$ the joint velocities. Additionally, slack variables \( \bm{\mathbf{s}} \in \mathbb{R}^{s} \) allow for geometric primitive flexibility when strict feasibility is momentarily unachievable. \( \mathbf{lb}_v, \, \mathbf{lb}_a \) and \( \mathbf{ub}_v, \, \mathbf{ub}_a \) define the lower and upper bounds for joint velocities, and accelerations, respectively, ensuring the robot operates within its feasible motion range.
The matrices \( \mathbf{A}_s \) and \( \mathbf{b}_s \) are defined as:
\begin{equation}
\mathbf{A}_s = 
\begin{bmatrix}
\bm{J}_{s \times r} & \mathbf{0}_{s \times s} \\
-\bm{J}_{s \times r} & \mathbf{0}_{s \times s} \\
\mathbf{0}_{s \times r} & \mathbf{I}_{{s \times s}}
\end{bmatrix}, 
\quad 
\mathbf{b}_s = 
\begin{bmatrix}
\gamma(\bm{\overline{\mathfrak{x}}^I_{i}}) - \bm{\Sigma_i} \\
- \gamma(\bm{\overline{\mathfrak{x}}^I_{i}}) + \bm{\Sigma_i} \\
\bm{\Sigma_{i}}
\end{bmatrix},
\end{equation}
where \( \gamma(\cdot) \) maps geometric primitives from dual quaternion space to cartesian space.
To prevent the solution from consistently reaching the slack constraint boundaries, the proposed QP framework assigns lower priority to slack minimization compared to other cost terms. The slack optimization is formulated as: $\bm{\mathcal{J}_s} = \operatorname{diag}(0_{r\times r}, \bm{I_{s \times s}})$, where $\operatorname{diag}(\cdot)$ is a block diagonal matrix.
The manipulability optimization $\bm{\mathbf{J}_\sigma}$ seeks to maximize the minimum singular value of the geometric jacobian, defined as $\mathbf{J_\sigma} = \left[ -\lambda \bar{\mathbf{J}}_\sigma, \, \mathbf{0}_{1 \times s} \right],$
where \( \bar{\mathbf{J}}_\sigma \) is the minimum singular value Jacobian with respect to \( \dot{\bm{\mathfrak{q}}} \).
To ensure effective obstacle avoidance, we implement an Artificial Potential Field (APF) approach, which introduces a repulsive force from obstacles acting on the end-effector. This modifies the original trajectory \( \bm{\overline{\mathfrak{x}}^I} \) to a new, adjusted trajectory \( \bm{\overline{\mathfrak{x}}^I_{\text{new}}} \), defined as:
\begin{equation}
\overline{\bm{\mathfrak{x}}^I_{\text{new}}} = r\left(\bm{\overline{\mathfrak{x}}^I}\right) + \frac{1}{2} \, \varepsilon \left( p\left(\bm{\overline{\mathfrak{x}}^I}\right) + \operatorname{Tr}\left( \bm{\mathcal{F}_{\text{att}}} + \bm{\mathcal{F}_{\text{rep}}} \right) \right) r\left(\bm{\overline{\mathfrak{x}}^I}\right),
\end{equation}
where \( \bm{\mathcal{F}_{\text{att}}} \) and \( \bm{\mathcal{F}_{\text{rep}}} \) denote the attractive and repulsive forces, respectively, as detailed in~\cite{khatib1986potential}. The operator \( r(\cdot) \) extracts the rotational component of a dual quaternion, \( p(\cdot) \) isolates its translational component, and \( \operatorname{Tr}(\cdot) \) maps a three-dimensional vector into quaternion space.
In summary, the hierarchical dual-stage motion architecture ensures safe and efficient trajectory planning during operations.

\section{Experimental Results}\label{sec:Expri}

This section focuses on the quantitative and experimental evaluation of the proposed framework. The planning algorithms were developed using the $DQ\_Robotics$ library~\cite{adorno2020dq} and implemented on two Franka Emika robot arm, each arm featuring seven degrees of freedom (7-DOF). We use the boost graph library\footnote{https://www.boost.org/} for graph construction. The system was controlled in a setup via the MuJoCo physics engine\footnote{https://mujoco.org/}. Utilizing a simulation environment enabled real-time interaction with dynamic external forces. The primary objectives of this setup were to streamline the evaluation process, enhance consistency and precision in handling external forces within planned scenarios, and ensure an equitable comparison with alternative methods.
In our method, to gather learning data from human demonstrations, we conducted ten trials of the robot grasping task on a real robot equipped with a modified impedance controller, continuously recording the end-effector position. Each demonstration was carried out under different initial and target robot configurations.
The planners were assessed across two force-intensive tasks: drilling and cutting. 
\begin{itemize}

\item \textbf{Task 1: Random Cutting}: The robot cuts randomly sized triangular shapes on a wooden board, with force fluctuating between 30 N and 60 N.

\item \textbf{Task 2: Random Drilling}: Five drilling operations were performed at random positions, perpendicular to the board, with an applied force of 15–20 N. Fig.~\ref{fig:tasks}(b), (c) shows two examples.

\item \textbf{Task 3: Combined Cutting and Drilling}: Three cutting and two drilling tasks with randomly generated force magnitudes.

\end{itemize}

\begin{figure*}
    \centering
    \begin{subfigure}[b]{0.93\linewidth}
        \centering
        \includegraphics[width=\linewidth, height=5.8cm]{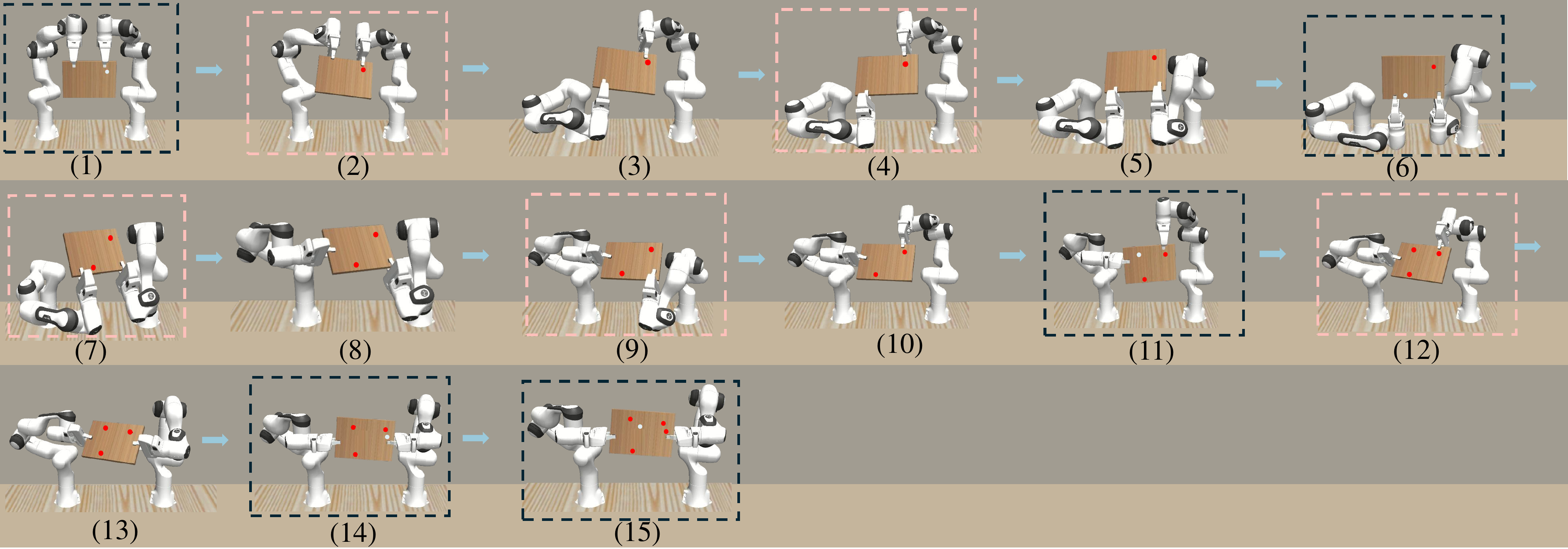}
        \caption{The minimal regrasp planner for a random-drilling task.}
        \label{fig:sequence1}
    \end{subfigure}
    
    \vspace{0.5cm} 

    \begin{subfigure}[b]{0.93\linewidth}
        \centering
        \includegraphics[width=\linewidth, height=3.8cm]{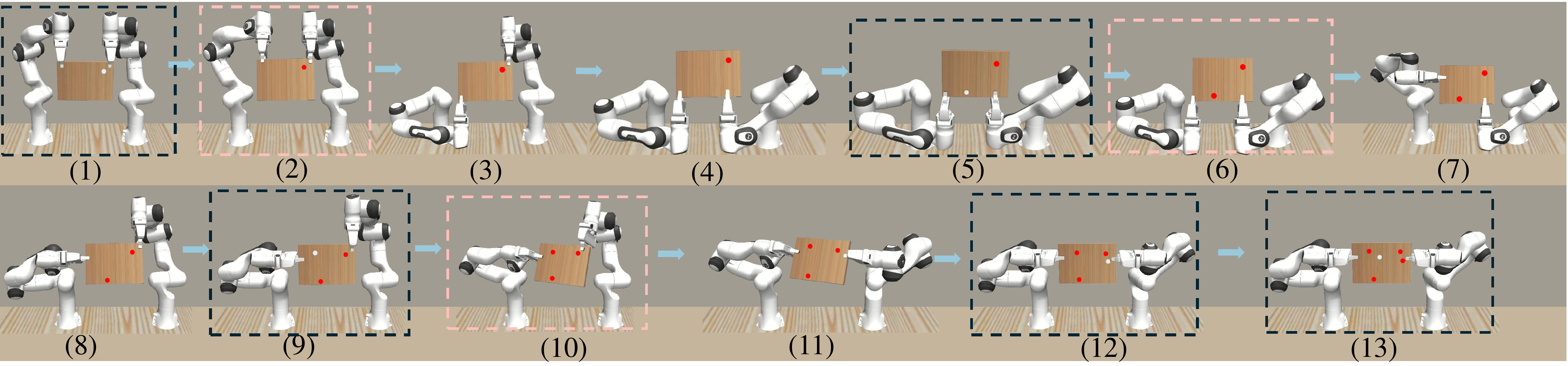}
        \caption{Our planner for a random-drilling task.}
        \label{fig:sequence2}
    \end{subfigure}
    
    \caption{Random-Drilling Task: The black dashed rectangle indicates force application configurations, the pink box represents transition intermediates, red dots mark previously applied points, and white dots denote active force points.}
    \label{fig:sequence_combined}
\end{figure*}

\subsection{Ablation Studies 1: Analysis of Motion Performance}

First, to evaluate the effectiveness of the Motion Performance Map for grasp configuration optimization (Section~\ref{map}) and the QP-based planning framework for motion performance optimization (Section~\ref{qp}), we conducted the \textbf{Task 1} (Fig.~\ref{fig:tasks}(\textcolor{black}{a})). Our method, which integrates motion performance components, was compared to a baseline version of our planner without motion performance considerations.
As shown in Fig.~\ref{fig:exp_sing}, while both methods satisfy external force constraints, our framework consistently achieves higher manipulability, measured by the minimum singular value of the geometric Jacobian. This improves adaptability to external disturbances and reduces singularity risks.
Additionally, we tracked the minimum joint distance from its limits throughout execution. Our method maintained a minimum gap of 0.557, whereas the baseline dropped to 0.187, demonstrating its effectiveness in preventing joint limit violations. The integration of the Motion Performance Map and manipulability constraints enhances stability and efficiency, ensuring safe and optimal performance under dynamic external forces.


\begin{table}[htbp]
\caption{Comparative performance statistics of different methods}
    \centering
    \resizebox{0.5\textwidth}{!}{ 
        \begin{tabular}{ccccccc}
            \toprule
            & \textcolor{black}{Execution} & \multicolumn{2}{c}{\text{EE Trajectory Length (m)}} & \multicolumn{2}{c}{\text{Min. Sing Value}} & \text{Min dis to} \\[-2pt]
            \cmidrule(lr){3-4} \cmidrule(lr){5-6}\\[-8pt]
            & \text{Time (s)} & \quad \text{left arm} & \text{right arm} & \text{left arm} & \text{right arm} & \text{joint limit} \\
            \midrule
            \text{Our} & $723.675$ & \quad 3.004 & 1.636 & 0.138 & 0.123 & 0.168 \\
            \text{Minimal Regrasp} & $815.772$ & \quad 3.673 & 2.161 & 0.025 & 0.076 & 0.031 \\
            \text{Random Sampling} & $1500.287$ & \quad 4.744 & 5.448 & 0.027 & 0.076 & 0.023 \\
            \text{Greedy} & $948.049$ & \quad 3.408 & 2.543 & 0.079 & 0.075 & 0.048 \\
            \bottomrule
        \end{tabular}
    }
\label{compared}
\end{table}

\subsection{Ablation Studies 2: Analysis of Planning Efficiency}

Second, to evaluate the efficiency of our proposed framework, we tested the Sampling Stable Intersections in Grasp Manifolds strategy (Section~\ref{grasp_sample}) integrated with the Dual-Stage Motion Architecture (Section~\ref{qp}) in \textbf{Task 2}, which involves discrete drilling forces randomly applied to a wooden board. This scenario, requiring frequent configuration transitions, serves as an ideal benchmark for computational efficiency.
We compared our planners against the method in~\cite{chen2020manipulation}, which randomly samples intermediate configurations and connects them using a modified BiRRT~\cite{rrt-connect} while minimizing regrasp operations. Across ten task instances, our approach achieved an average execution time of \(529.296 \pm 128.974\) s, significantly faster than the baseline’s \(728.994 \pm 122.257\) s. Additionally, our method reduced end-effector travel distances, with the left arm moving \(\bigl(1.888 \pm 0.245\bigr)\) m and the right arm \(\bigl(2.483 \pm 0.233\bigr)\) m, compared to the baseline’s \(\bigl(2.789 \pm 0.845\bigr)\) m and \(\bigl(2.941 \pm 0.286\bigr)\) m, respectively. By integrating imitation learning for global path planning and APF for collision avoidance, our method outperforms RRT-based approaches in real-time execution.
In addition, this efficiency gain also stems from the Grasp Manifolds Layer, which reduces unnecessary intermediate configurations. As shown in Fig.~\ref{fig:sequence_combined}, our method completes all transitions in just three intermediate steps, whereas the baseline requires five, as highlighted in the pink rectangle.
Moreover, our approach generates intermediate configurations along the shortest-cost path, as shown in Fig.~\ref{fig:sequence_combined} (a-2 vs. b-2, a-7 vs. b-6, and a-12 vs. b-10), minimizing board movement and achieving equilibrium more efficiently. 

In conclusion, our framework delivers superior planning efficiency by reducing computational overhead, lowering transition costs, and enhancing real-time adaptability in dynamic grasping scenarios.

\subsection{Comparison Experiment}
Finally, we compared our approach with three planners: (i) the Minimal Regrasp Planner~\cite{chen2019manipulation}, (ii) a Random Sampling Planner, and (iii) a Greedy Planner. The Random Sampling Planner selects random configurations until a feasible one is found for the initial force. For subsequent forces, it checks stability; if unstable, it resamples. Unlike the Minimal Regrasp Planner, which globally minimizes grasp transitions across all forces, the Greedy Planner optimizes locally, minimizing grasp transitions at each discrete force change. We conducted \textbf{Task 3} experiments to verify feasibility under a complex hybrid external force variation task, as illustrated in Fig.~\ref{fig:tasks}(\textcolor{black}{d}). 
According to the results summarized in Table.~\ref{compared}, our algorithm achieves lower running time and shorter travel distance compared to the other planners. It further offers substantial improvements in manipulability (Min. Sing Value) and maintains a higher distance to joint limits. Thus, the comprehensive planning framework presented here provides excellent efficiency and robust motion performance.

\section{Conclusion and future work}  

This paper proposed an imitation-guided bimanual planning framework that enhanced stability and adaptability under varying external forces. By optimizing grasp transitions and leveraging a dual-stage motion architecture combining imitation learning with quadratic programming, it achieved efficient, real-time trajectory generation with minimal regrasping. Experiments demonstrated superior performance in force-intensive tasks, with shorter execution times and reduced trajectory lengths. Future work will extend to multi-contact scenarios and integrate reinforcement learning for adaptive force control.







\bibliographystyle{IEEEtran}
\bibliography{references}
\balance

\end{document}